\documentclass[lettersize,journal]{IEEEtran}
\usepackage{amsmath,amsfonts}
\usepackage{algorithmic}
\usepackage{algorithm}
\usepackage{array}
\usepackage[caption=false,font=normalsize,labelfont=sf,textfont=sf]{subfig}
\DeclareUnicodeCharacter{00A0}{ }
\usepackage{textcomp}
\usepackage{stfloats}
\usepackage{url}
\usepackage[bookmarks=false]{hyperref}
\usepackage{verbatim}
\usepackage{graphicx}
\usepackage{multirow}
\usepackage{booktabs}
\usepackage{cite}
\begin{document}

\title{A CNN-BiLSTM Model with Attention Mechanism for Earthquake Prediction}

\author{Parisa Kavianpour$^1$, Mohammadreza Kavianpour$^2$, Ehsan jahani$^1$, Amin Ramezani$^2$ \\
$^1$Department of Civil Engineering, University of Mazandaran, Babolsar, Iran\\
$^2$Department of Electrical and Computer Engineering, Tarbiat Modares University, Tehran, Iran\\
E-mail address: kavianpour@modares.ac.ir
\thanks{\textbf{Copyright may be transferred without notice, after which this version may no longer be accessible.}}

}
\maketitle
\begin{abstract} 
 Earthquakes, as natural phenomena, have continuously caused damage and loss of human life historically. Earthquake prediction is an essential aspect of any society's plans and can increase public preparedness and reduce damage to a great extent. 
Nevertheless, due to the stochastic character of earthquakes and the challenge of achieving an efficient and dependable model for earthquake prediction, efforts have been insufficient thus far, and new methods are required to solve this problem. Aware of these issues, this paper proposes a novel prediction method based on attention mechanism (AM), convolution neural network (CNN), and bi-directional long short-term memory (BiLSTM) models, which can predict the number and maximum magnitude of earthquakes in each area of mainland China-based on the earthquake catalog of the region.
This model takes advantage of LSTM and CNN with an attention mechanism to better focus on effective earthquake characteristics and produce more accurate predictions.
Firstly, the zero-order hold technique is applied as pre-processing on earthquake data, making the model's input data more proper. Secondly, to effectively use spatial information and reduce dimensions of input data, the CNN is used to capture the spatial dependencies between earthquake data. Thirdly, the Bi-LSTM layer is employed to capture the temporal dependencies. Fourthly, the AM layer is introduced to highlight its important features to achieve better prediction performance.
The results show that the proposed method has better performance and generalize ability than other prediction methods.

\end{abstract}

\begin{IEEEkeywords}
Earthquake prediction, convolution neural network, LSTM, time series prediction, attention mechanism
\end{IEEEkeywords}

\section{Introduction}
 Earthquakes are a type of natural catastrophe that can be extremely devastating. They are unexpected events and often occur without any prior warning, imposing human and financial losses to society \cite{alarifi2012earthquakes}. In addition, it can cause other various disasters such as flood and avalanche \citeonline{cui2011wenchuan}, tsunamis \cite{jain2019proficient} and debris collapse \cite{cui2013risk}, and also various consequences for the ecosystem like soil liquefaction \cite{verdugo2015liquefaction} and fault rupture \cite{bray2001developing}.  According to the high rate of damage and mortality \cite{bilham2009seismic,ambraseys2005history }, also the primary and secondary consequences of earthquakes \cite{jia2017earthquake}, researchers have conducted extensive research for many years to increase people's awareness of the phenomenon of earthquakes. They have proposed various solutions for earthquake prediction \cite{sobolev2015methodology}. Timely and reasonable prediction is critical to prevent or reduce the destructive effects of this phenomenon, and it aids society in developing more accurate scenarios of the crisis process and taking measures to manage it. In other words, such a prediction improves the level of readiness and speed of action in crisis control in the community. An efficient prediction specifies the magnitude and geographical location of a future earthquake over a period of time \cite{kail2021recurrent}. It can save plenty lives and prevent severe economic losses. However, many methods have been proposed based on the features and parameters affecting the earthquake, but only a few of these works have achieved accurate predictions because earthquakes are a stochastic and complex phenomenon involving a wide range of parameters that are difficult to analyze \cite{otari2012review}. Mathematical modeling, precursor signal analysis, shallow machine learning (ML) techniques, and deep learning (DL) algorithms are some of the most popular methods in earthquake prediction research. Table \ref{tab:RELATED} displays the works related to earthquake prediction regarding to four categories.
\\

In the mathematical modeling attempts to formulate earthquake prediction using various statistical and mathematical methods \cite{boucouvalas2015modified}. For instance, Marisa et al.\cite{marisa2019earthquake} applied a Poisson hidden Markov model (PHMM) to predict the probability of an earthquake in Sumatra island. Dehghani and Fadaee \cite{dehghani2020probabilistic} proposed a bivariate lognormal distribution-based probabilistic prediction method.  Its parameters are adjusted using the maximum likelihood method. They were able to predict the probability of an earthquake in Tehran with a magnitude of 6.6 to 6.8 in the next 10 to 15 years. Kannan \cite{kannan2014innovative} utilizes the hypothesis of the spatial connection to predicting the location of earthquakes in six major seismic regions. Spatial connections theory states that earthquakes within a fault zone are related to each other. Statistical approaches typically need prior assumptions such as stationarity and linear correlation between data, while earthquake data are non-stationarity. Therefore, the applying of these methods for earthquake prediction frequently does not yield good results \cite{aggarwal2020statistical}.
\\

Another branch of research focuses on precursor signals analysis. Radon gas emissions \cite{woith2015radon}, unusual animal behavior \cite{wikelski2020potential}, electromagnetic signals \cite{uyeda2009short}, ionospheric analysis \cite{li2013statistical}, and other anomalous phenomena that may indicate an impending earthquake are examples of precursors. Uyeda et al. \cite{uyeda2009short} provided short-term earthquake predictions based on electromagnetic signals analyses. Li and Parrot \cite{li2013statistical} claimed that ionospheric density variations could be used as a seismic precursor. In another study, Wikelski et al. \cite{wikelski2020potential} examined the behaviors activities of farm animals in order to predict the short-term earthquake. The majority of the mentioned techniques are reliant on the occurrence of particular precursors. Predictions based on precursors rarely produce satisfactory results because they may occur without any further earthquake event and are difficult to detect. 
\\

The next category of research refers to the application of machine learning approaches in predicting earthquake. These approaches are data-driven, non-parametric, and require fewer a priori assumptions. In their study, Murwantara et al. \cite{murwantara2020comparison} used three different neural networks, including multinomial logistic regression (LR), support vector machine (SVM), and Naive Bayes (NB), to predict the magnitude, location, and depth of earthquakes in Indonesia. They confirmed that the SVM algorithm performs better than the other two methods in earthquake prediction.  khalil et al. \cite{khalil2021integrated} developed an earthquake prediction approach along the Chaman fault in Baluchistan based on a hybrid neural network (HNN) and SVM approach. Lin. \cite{lin2020researching} proposed probability backpropagation neural network (BPNN) for prediction of earthquake Probabilistic in Taiwan. ML approaches have a limited ability to learn nonlinear and complex relationships of earthquake data. In addition, they only can extract shallow features in the dataset and almost need complex feature engineering operations. 
\\

DL algorithms have recently made significant progress in solving a wide range of earthquake prediction problem. Because these models include multiple hidden layers and densely connected many neurons, they have a high generalization power, which has significantly increased their learning ability compared to shallow network. Many approaches based on DL have been expanded for earthquake prediction, such as convolutional neural networks (CNN) \cite{huang2018large} and long short-term memory (LSTM) networks \cite{bhandarkar2019earthquake}. Huang et al.\cite{huang2018large} suggested the CNN model in their research to predict the magnitude of the major earthquake in Taiwan based on image data. Jozinović et al. \cite{jozinovic2020rapid} offers a CNN-based approach for predicting earthquake ground shaking intensity measurements in Italy.  Li et al. \cite{li2020dlep} introduced a Deep Learning model for Earthquake Prediction called DLEP by merging explicit and implicit earthquake characteristics. In DLEP, they employ eight precursory pattern-based indications as the explicit features, and utilize a CNN to extract implicit features.
Bhandarkar et al. \cite{bhandarkar2019earthquake} examined the trend of upcoming earthquake using the LSTM network. Wang et al. \cite{wang2017earthquake} employed an LSTM network to learn the spatial-temporal correlation between earthquakes in different areas and make predictions based on it. Al Banna et al. \cite{al2021earthquake} have suggested an attention-based bi-directional LSTM architecture for earthquake prediction in Bangladesh in the coming month.
\\

In earthquake prediction, it is critical to collect both the spatial and temporal information of the earthquake data at the same time. Nevertheless, the above studies frequently ignore the spatial or temporal characteristics of earthquake data, Which causes some hidden features in the data not to be extracted well. Combining CNN and LSTM methods can improve forecasting performance by combining their benefits. Nicolis et al. \cite{nicolis2021prediction} proposed a hybrid CNN-LSTM approach for earthquake prediction in Chile, and they utilized geographic images related to seismic data as input. Their study's goal have been to predict the location and intensity of seismic events. Mousavi and Beroza \cite{mousavi2020machine} introduced a CNN-RNN network for earthquake magnitude estimation. They examined the time-frequency characteristics of the dominant phases in an earthquake signal from three component data recorded on individual stations in Northern California.
 Since earthquake data is a time series, it has a stronger prediction potential when both past and future features are analyzed simultaneously.\\
 
 By examining studies in the field of earthquake prediction, it is obvious that for more accurate earthquake prediction, a method should be proposed that, in addition to simultaneously extracts the temporal and spatial features of the input, the influence of characteristics extracted from input variables to output to have the smallest amount of prediction error. In this study, a novel hybrid CNN-BiLSTM method based on attention mechanism (AM) is proposed to predict the maximum magnitude of earthquakes and the number of earthquakes in a specified time period in Mainland China. The proposed method merges the benefits of CNN and LSTM, and the AM can assess to dynamically the significance of each feature in the sample input.
  The spatial features are extracted by the CNN layer, the BiLSTM layer is capable to process the extracted feature in order to learn the long-term dependency of the data , and the attention layer assigns high weights on important features. Also, appropriate preprocessing of input data affects the performance of a model. So, we utilize Zero-order hold techniques to preprocess the data to improve the results and generalization ability of the model.


\begin{table*}[t]
\caption{Related Works on Earthquake Prediction}
\label{tab:RELATED}
\resizebox{\textwidth}{!}{%
\begin{tabular}{ccccc}
\hline
\textbf{Category}                                & \textbf{Authors}                                                                & \textbf{Year}            & \textbf{Method}                     & \textbf{Description}                                                                                                                                    \\ \hline
\multirow{5}{*}{Mathematical method}             & S. Kannan. \cite{kannan2014innovative}                         & 2014                     & Poisson distribution                & \begin{tabular}[c]{@{}c@{}}Predicting the location of the next earthquake \\ according to the theory of spatial connections\end{tabular}                \\ \cline{2-5} 
                                                 & Marisa et al. \cite{marisa2019earthquake}                      & 2019                     & PHMM                                & \begin{tabular}[c]{@{}c@{}}applied a PHMM model to predict the probability\\  of an earthquake in Sumatra Island\end{tabular}                           \\ \cline{2-5} 
                                                 & H. Dehghani and M. J. Fadaee. \cite{dehghani2020probabilistic} & 2020                     & BLD                                 & \begin{tabular}[c]{@{}c@{}}Predicting the time and magnitude of a future  \\ earthquake in Tehran using a model probabilistic\end{tabular}              \\ \hline
\multirow{5}{*}{Precursor method}                & S. Uyeda et al. \cite{uyeda2009short}                          & 2009                     & statistical analysis                & \begin{tabular}[c]{@{}c@{}}Provide short-term earthquake predictions based \\ on electromagnetic signal analysis\end{tabular}                           \\ \cline{2-5} 
                                                 & M. Li and M. Parrot. \cite{li2013statistical}                  & 2013                     & statistical analysis                & \begin{tabular}[c]{@{}c@{}}earthquake prediction based on ionospheric density\\  variations\end{tabular}                                                \\ \cline{2-5} 
                                                 & M. Wikelski et al. \cite{wikelski2020potential}                & 2020                     & statistical analysis                & \begin{tabular}[c]{@{}c@{}}Investigate the behaviors and activities of farm \\ animals in order to predict short-term earthquakes\end{tabular}          \\ \hline
\multirow{5}{*}{Shallow machine learning method} & I. M. Murwantara et al. \cite{murwantara2020comparison}        & 2020                     & Multinomial LR, SVM and Naive Bayes & \begin{tabular}[c]{@{}c@{}}Medium and long-term earthquake prediction \\ in Indonesia using 30 years of historical data\end{tabular}                    \\ \cline{2-5} 
                                                 & JW. Lin. \cite{lin2020researching}                             & \multicolumn{1}{l}{2020} & BPNN                                & \begin{tabular}[c]{@{}c@{}}Probabilistic earthquake prediction in Taiwan\\  using 25 years of historical data\end{tabular}                              \\ \cline{2-5} 
                                                 & U. Khalil et al. \cite{khalil2021integrated}                   & \multicolumn{1}{l}{2021} & HNN-SVM                             & \begin{tabular}[c]{@{}c@{}}Earthquake prediction along Chaman fault \\ of Baluchistan based on seismic indicators\end{tabular}                          \\ \hline
\multirow{15}{*}{Deep learning method}            & Q.Wang et al. \cite{wang2017earthquake}                        & \multicolumn{1}{l}{2017} & LSTM                                & \begin{tabular}[c]{@{}c@{}}Earthquake prediction by learning the \\ spatial-temporal correlation between earthquakes \\ in different areas\end{tabular} \\ \cline{2-5} 
                                                 & JP. Huang et al. \cite{huang2018large}                         & \multicolumn{1}{l}{2018} & CNN                                 & \begin{tabular}[c]{@{}c@{}}Major earthquake prediction in Taiwan\\  based on image data\end{tabular}                                                    \\ \cline{2-5} 
                                                 & T. Bhandarkart et al. \cite{bhandarkar2019earthquake}          & 2019                     & LSTM                                & \begin{tabular}[c]{@{}c@{}}Predict future earthquake trends using \\ earthquake historical data\end{tabular}                                            \\ \cline{2-5} 
                                                 & R. Li et al. \cite{li2020dlep}                                 & 2020                     & CNN                                 & \begin{tabular}[c]{@{}c@{}}Earthquake prediction by combining \\ explicit and implicit earthquake features\end{tabular}                                 \\ \cline{2-5} 
                                                 & D. Jozinović. \cite{jozinovic2020rapid}                        & \multicolumn{1}{l}{2020} & CNN                                 & \begin{tabular}[c]{@{}c@{}}Prediction of earthquake ground shaking \\ intensity using raw waveform in Italy data\end{tabular}                           \\ \cline{2-5} 
                                                 & Mousavi et al. \cite{mousavi2020machine}                       & \multicolumn{1}{l}{2020} & CNN-BiLSTM                          & \begin{tabular}[c]{@{}c@{}}Earthquake magnitude estimation in \\ Northern California based on earthquake signal\end{tabular}                            \\ \cline{2-5} 
                                                 & Al Banna et al. \cite{al2021earthquake}                        & \multicolumn{1}{l}{2021} & BiLSTM-AM                           & \begin{tabular}[c]{@{}c@{}}Earthquake prediction in Bangladesh in the\\  coming month based on eight seismic indicators\end{tabular}                    \\ \cline{2-5} 
                                                 & O. Nicolis et al. \cite{nicolis2021prediction}                 & 2021                     & CNN-LSTM                            & \begin{tabular}[c]{@{}c@{}}Earthquake prediction in Chile based \\ on geographic images\end{tabular}                                                    \\ \hline
\end{tabular}%
}
\end{table*}


Our basic contributions in this research can be summed up as follows:
\begin{enumerate}

\item A novel hybrid CNN-BiLSTM-AM method for earthquake prediction is suggested, which predicts the maximum magnitude and number of earthquakes in one month period.
\item Due to the nature of the earthquake data, the CNN and BiLSTM-based networks is applied to fully extract the spatial-temporal characteristics of earthquake data. AM is also employed to focus on the important characteristics that have a strong correlation with earthquake prediction results.
\item We have divided Mainland China into nine smaller sub-regions to determine the range of location of the next earthquake. The ZOH technique for data preprocessing has been used to reduce the effect of data's zero-value on the network training process and reduce prediction error. The simulation results demonstrate the proposed method's efficiency compared to other methods.
\end{enumerate}
The rest of this paper is organized as follows. Section II is shown the preliminary knowledge. Section III is described the proposed method.  experiment and compared method are provided in section IV. Section V is presented introduces our case stud and results analysis. Section VI concludes this study.

\section{PRELIMINARIES} \label{sec2}

\subsection{Covolutional Neural Network}
CNN has successfully been applied in image processing, face recognition, and time series analysis \cite{li2020prediction}. The CNN architecture is created by stacking three main layers: convolution, pooling, and fully connected (FC). There is a set of learnable filters in each convolution layer whose goal is to extract local characteristics from the input matrix automatically. Filters perform convolution operations based on two important ideas, namely weight sharing and local connection, which can help to reduce the complexity of the computational burden and boost model performance \cite{lu2020cnn}. The pooling layer follows the convolution layer and performs the downsampling operation. One distinguishing characteristic of pooling layer is that it reduces the feature map's dimensionality and avoids over-fitting. Usually, the FC layers are employed in the last layers of CNN architecture to learn the nonlinear combination of features extracted by the convolution layer to create the final output \cite{barzegar2020short}.

\subsection{Bidirectional LSTM}
\begin{figure}[t]

    \centering
    \includegraphics[width=\columnwidth ]{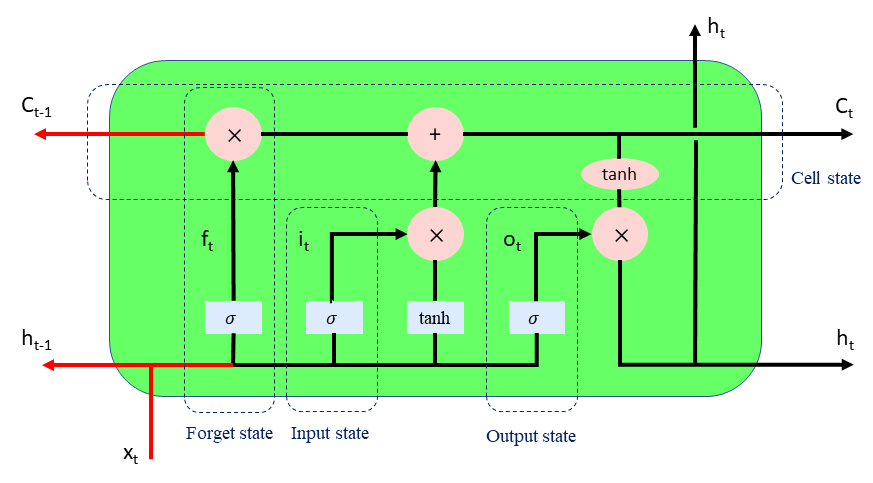}
    \caption{schematic diagram of bidirectional LSTM}
    \label{bilstm}
    \end{figure}

The RNN model, which is used to analyze time series data, contains a return loop to use prior information efficiently. However, RNN has limitations on memory and information storage. It cannot well learn long-term dependencies, and lead to vanishing gradient \cite{miao2020short}. As a result, the LSTM was designed to address the shortcomings of RNN.  
The LSTM structure is based on using memory cells to remember long-term historical information and regulating this through a gate mechanism.
A common LSTM unit has three types of gates: input gate ${i_t}$, forget gate ${f_t}$, and output gate ${o_t}$. These three gates are shown in Fig. \ref{bilstm}. In each gate, controlling the state of memory cells is performed through point-wise multiplication and sigmoid function operations. Input data ${x_t}$ at the current state and output ${h_{t - 1}}$ from the hidden state of the previous layer enter all gates. The forget gate decides what information should be ignored or kept.
Information from the current input ${x_t}$ and from the previously hidden state ${h_{t - 1}}$ are transitioned by the sigmoid function. Thus, the forget gate's output value is between zero and one. If the value is near to zero, it indicates that the information will be discarded.  Otherwise, the closer to one, the more information will be kept. The formula of the forget gate is calculated as follows:
\begin{equation}
\begin{aligned}
  {f_t} = \sigma \left( {{W_f}.\left[ {{h_{t - 1}},{x_t}} \right] + {b_f}} \right)
\end{aligned}
\end{equation}
Where $\sigma $ is the sigmoid activation function, and $W$ and $b$ denotes the weight and bias of any gate unit, respectively.
 The the current input ${x_t}$ and the previously hidden state ${h_{t - 1}}$ are fed into the sigmoid function. By transforming the values from zero to one, the input gate decides which information are updated. Among them, zero denotes unimportant and one denotes importance. The input gate is formulated as:
\begin{equation}
\begin{aligned}
  {i_t} = \sigma \left( {{W_i}.\left[ {{h_{t - 1}},{x_t}} \right] + {b_i}} \right)
\end{aligned}
\end{equation}
Then the current input ${x_t}$ and hidden state ${h_{t - 1}}$ are fed to the $tanh$ function.
Now, the cell state ${{\hat C}_t}$ is calculated and the new value is updated to the cell state.  

  \begin{equation}
\begin{aligned}
 {{\hat C}_t} = \tanh ({W_c}.\left[ {{h_{t - 1}},{x_t}} \right] + {b_c})
\end{aligned}
\end{equation}

  \begin{equation}
\begin{aligned}
  {C_t} = {f_t} \odot {C_{t - 1}} + {i_t} \odot {{\hat C}_t}
\end{aligned}
\end{equation} 
Where, $tanh$ is the hyperbolic tangent activation function. $\odot $ is the dot multiplication operation, and ${C_t}$ is the new memory cell. Finally, the output gate selects the next hidden state. The new memory cell ${C_t}$ and new hidden state ${h_t}$ are then passed to the next time step.
  \begin{equation}
\begin{aligned}
  {o_t} = \sigma \left( {{W_o}.\left[ {{h_{t - 1}},{p_t}} \right] + {b_o}} \right)
\end{aligned}
\end{equation} 
 \begin{equation}
\begin{aligned}
{h_t} = {o_t}\odot\tanh ({c_t})
  \end{aligned}
\end{equation}

 A single LSTM usually  processes information from just one forward direction. In other words, it can only use prior information. In contrast, the structure of BiLSTM is such that it has two layers of LSTM, one at the forward and the other at the backward. Fig. \ref{bilstm} depicts the schematic diagram of  BiLSTM. The forward LSTM can receive the input sequence's past data information, whereas the reverse LSTM can obtain the input sequence's future data information, then the output in both hidden layers is combined.
The hidden state ${h_t}$ of Bi-LSTM at current time ${t}$ contains both of forward ${{\vec h}_t}$ and backward ${{\overset{\lower0.5em\hbox{$\smash{\scriptscriptstyle\leftarrow}$}}{h} }_t}$ :
\begin{equation}
\begin{aligned}
{h_t} = {{\vec h}_t} \oplus {{\overset{\lower0.5em\hbox{$\smash{\scriptscriptstyle\leftarrow}$}}{h} }_t}
\end{aligned}
\end{equation} 

Where $ \oplus $ denotes the summation by component, used to sum the forward and backward output components. Bi-LSTM produces better efficiency than LSTM and RNN  because it can use both preceding and subsequent information.
\\

\subsection{Attention Mechanism}
AM is an appropriate idea for upgrading the importance of crucial information inspired by the human visual system. When human vision observes anything in the environment, it does not usually witness a scene from beginning to end, but rather focuses on a specific section as needed. Based on, the AM selectively focuses on some of the more influential information, dismisses unnecessary information, and boosts desirable information. AM is commonly used in various fields such as image captioning \cite{chu2020automatic}, machine translation \cite{zhang2020neural}, earthquake prediction \cite{al2021attention}. AM acts based on weight allocation,  determining the most effective information by distributing higher weights. As a result, it has a positive optimization impact on traditional models. The nature of the attention function can be defined as a mapping from a Query to a sequence of Key-Value pairs. And, as seen in Fig. \ref{am}, calculating attention involves three phases. In the first phase, the similarity or correlation between the Query and each Key is calculated as follows:  
 \begin{figure}[t]
    \centering
    \includegraphics[width=\columnwidth ]{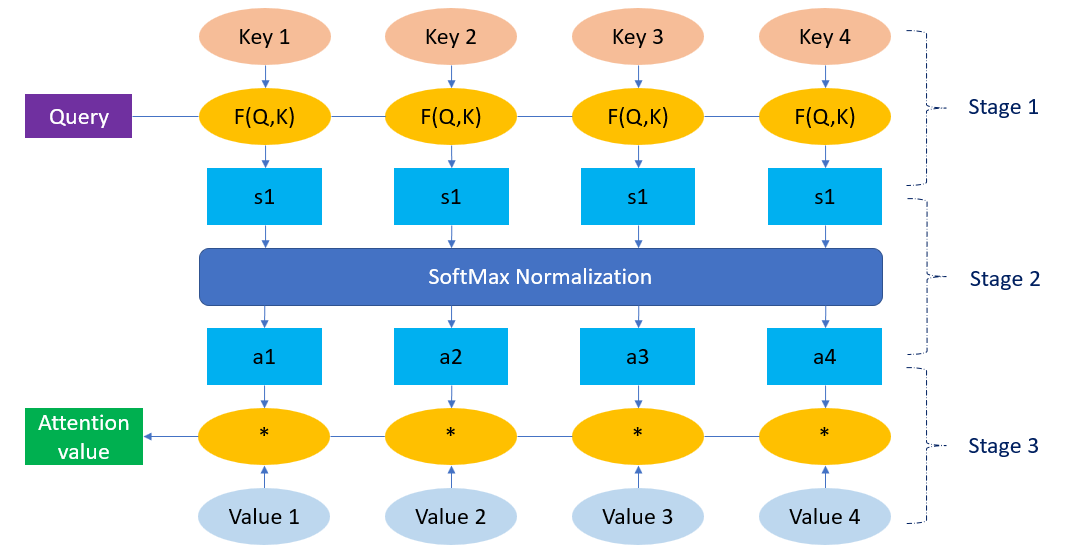}
    \caption{The step in determining AM}
    \label{am}
    \end{figure}

 \begin{equation}
     \begin{aligned}
      {s_t} = \tanh ({W_h}{h_t} + {b_h})
     \end{aligned}
 \end{equation}
 
where, ${s_t}$ is attention score. $W_h$, $b_h$ are the weight and bias of AM, respectively. $h_t$ is the input vector. In the second phase, The score obtained of the first stage is normalized, and the softmax function is utilized to convert the attention score as given in formula:
 \begin{equation}
     \begin{aligned}
      {a_t} = \frac{{\exp (s_t)}}{{\sum\limits_t {\exp (s_t)} }}
     \end{aligned}
 \end{equation}

 regarding to the weight coefficient, the final attention value is obtained by weighted summation of value as shown in formula :
 \begin{equation}
     \begin{aligned}
     s = \sum\limits_t {{a_t}{h_t}} 
     \end{aligned}
 \end{equation}
 
 \begin{figure*}[t]

    \centering
    \includegraphics[width=\textwidth]{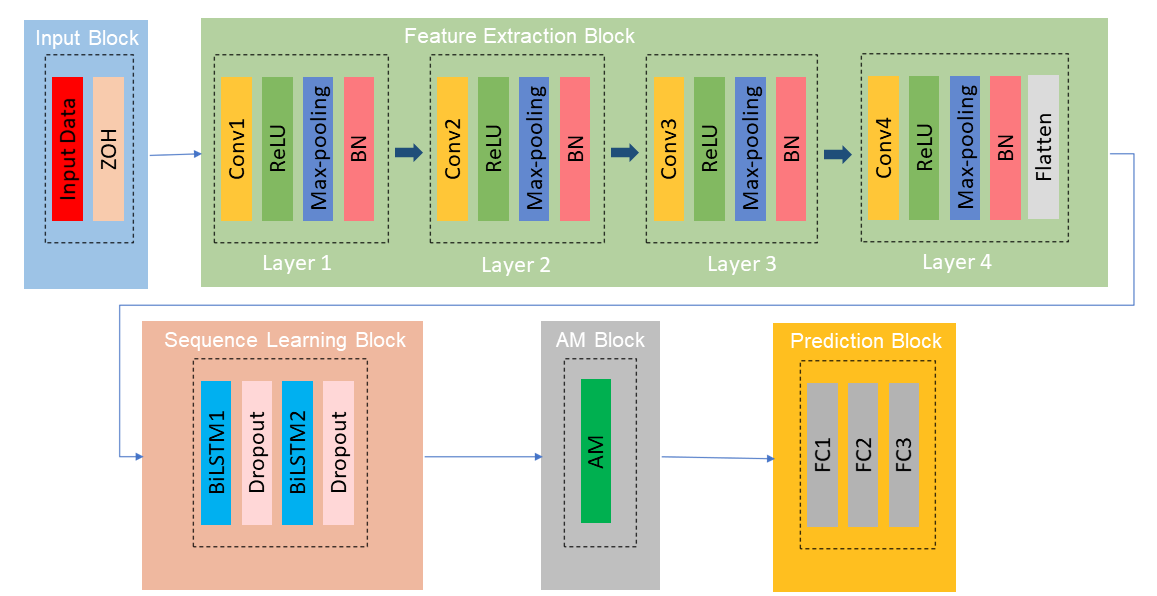}
    \caption{The architecture of proposed method }
    \label{proposed}
    \end{figure*}

  AM usually is used after CNN and RNN networks to focus on the features that significantly influence output variables, hence increasing the model's performance.

\section{THE FRAMEWORK OF THE PROPOSED METHOD} \label{sec2}
    
\subsection{CNN-BiLSTM-AM}
 In this section, we describe the proposed method architecture of earthquake prediction and its main components. To extract features more effectively and enhance prediction performance, we integrate CNN, BiLSTM, and AM into a single framework and suggest a novel CNN-BiLSTM-AM earthquake prediction method.  As shown in Fig. \ref{proposed}, the proposed method consists of five basic blocks: input block, feature extraction block, sequence learning block, attention block, and the prediction block. In the feature extraction block, CNN is utilized to extract spatial features from the input data, and these recovered spatial features are sent into the BiLSTM network as input. The BiLSTM is used to learn long-term temporal information in the sequence learning block, and the results are fed as input to the AM layer. In attention block, AM through the model feature input, assigns various weights, highlighting the effect of the more significant component, and assists the model in making more correct decisions. Finally, fully connected layers and output layer are stacked in the prediction block to carry out the final prediction. Each of these parts contains trainable parameters such as filter size, lose function, number of neurons, and kernels that adjusting optimally of these parameters can reduce the prediction error of the proposed method. The structural details of the proposed method are given in Table \ref{tabl11}. In the following, each block of the proposed method are described in detail: 
\\
\textbf{1) Input block:} The input block contains the earthquake information in the per month, which are preprocessed through ZOH technique. The ZOH technique helps to reduce the effect of data's zero-value on the network training process and improve prediction performance.
\\
\textbf{2) Feature extraction block:} The feature extraction block is obtained by a one-dimensional CNN with nine layers, including four convolutional layers, four pooling layers, and one flatten layer. The filter size is wide selected in the first convolution layer of convolution, unlike the subsequent layers. When compared to small kernels, this structure is superior at damping high-frequency signals. Stacking several convolutional and pooling layers allows higher-level features to be extracted from the input, which helps represent the input data better. The Max-pooling layer is implemented after each convolution layer to reduce the dimensions and parameters within the network. In the feature extraction block, rectified linear unit (ReLU) is used as the activation function to avoid gradient vanishing or explosion problems while enhancing the convergence rate. Following each convolution layer, a batch normalization (BN) algorithm is employed as an effective regularization strategy. In addition to having a regularizing effect, it can reduce the shift of internal covariate, better the network's training performance, and increase the generalization capability of the network. BN is a feature normalization method in a layer-by-layer manner that is applied to accelerate the speed of the training process. Features in each layer are first normalized to the standard distribution and are then regulated to the ideal distributions. BN method is calculated as follows:

\begin{equation}
\begin{aligned}
 \mu  = \frac{1}{{{N_{batch}}}}\sum\nolimits_{i = 1}^N {{x_i}}  \hfill
\end{aligned}
\end{equation} 

\begin{equation}
\begin{aligned}
{\sigma ^2} = \frac{1}{{{N_{batch}}}}{\sum\nolimits_{i = 1}^N {\left( {{x_i} - \mu } \right)} ^2} \hfill
\end{aligned}
\end{equation} 

\begin{equation}
\begin{aligned}
{{\overset{\lower0.5em\hbox{$\smash{\scriptscriptstyle\frown}$}}{x} }_i} = \frac{{{x_i} - \mu }}{{\sqrt {{\sigma ^2} + \varepsilon } }} \hfill
\end{aligned}
\end{equation} 

\begin{equation}
\begin{aligned}
  {y_i} = \gamma {{\overset{\lower0.5em\hbox{$\smash{\scriptscriptstyle\frown}$}}{x} }_i} + \beta  \hfill \\ 
\end{aligned}
\end{equation} 

Where $N_{batch}$ defines the size of mini-batch, ${x_i}$ and ${y_i}$ are the input and output of the ${i^{th}}$ observation value in the mini-batch. $\mu$ denotes the mean value of the mini-batch sample. $\gamma$ is the standard deviation of the mini-batch sample. $\varepsilon $ is a constant close to zero to ensure numerical stability, $\gamma $ is a scaling parameter, and $\beta$ is a bias parameter.  
The padding type is chosen same to ensure that no features are lost during convolution operations. Since the input data of the BiLSTM layer are required to be a one-dimensional array, the multidimensional output data of the Convolution layer needs to be flattened into one-dimensional data by the flatten layer.
\\
\textbf{3) Sequence learning block:} Sequence learning block aids in the learning of the temporal patterns of properties extracted via the feature extraction block. The sequence learning segment contains two BiLSTM layers and two dropout layers. After each BiLSTM layers, the dropout technique is used to prevent the over-fitting issue. Dropout means that instead of training all the neurons in the network, only some of them are randomly selected and trained. In short, a certain percentage of neurons in each iteration train take zero output and are inactivated. It drives the network to learn more effective features and boosts the model's generalization capacity.  
\\
\textbf{4) Attention block:} In the attention block, an attention layer is added to the end of the sequence learning block that highlights the factors that are more influential on prediction results and improves the prediction accuracy.
\\
\textbf{5) Prediction block:} The prediction block is made up of two fully connected layers as well as an output layer. The fully connected layer performs a series of nonlinear transformations on the values of the features obtained by the attention block. In the end, the final prediction results are generated.

\begin{table}[]
\caption{Structures of Proposed Method}
\label{tabl11}
\resizebox{\columnwidth}{!}{%
\begin{tabular}{ccc}
\hline
\textbf{Block}                                                                     & \textbf{Layer} & \textbf{\begin{tabular}[c]{@{}c@{}}Number/ size/ stride of kernels \\ or number of neurons\end{tabular}} \\ \hline
\multirow{9}{*}{Feature extraction block}                                          & Conv1          & 16*5*1                                                                                                   \\
                                                                                   & Max-pooling    & 16/2/1                                                                                                   \\
                                                                                   & Conv2          & 32*3*1                                                                                                   \\
                                                                                   & Max-pooling    & 32/2/1                                                                                                   \\
                                                                                   & Conv3          & 64*3*1                                                                                                   \\
                                                                                   & Max-pooling    & 64/2/1                                                                                                   \\
                                                                                   & Conv4          & 128*3*1                                                                                                  \\
                                                                                   & Max-pooling    & 128/2/1                                                                                                  \\
                                                                                   & Flatten        & -                                                                                                        \\ \hline
\multirow{2}{*}{\begin{tabular}[c]{@{}c@{}}Sequence learning\\ block\end{tabular}} & BiSTM1         & 128                                                                                                      \\
                                                                                   & BiLSTM2        & 64                                                                                                       \\ \hline
Attention block                                                                    & Attention      & -                                                                                                        \\ \hline
\multirow{3}{*}{Prediction block}                                                  & FC1            & 32                                                                                                       \\
                                                                                   & FC2            & 10                                                                                                       \\
                                                                                   & FC3            & 1                                                                                                        \\ \hline
\end{tabular}%
}
\end{table}

\subsection{The overall procedure of the proposed CNN-BiLSTM-AM method}  
\begin{figure}[t]

    \centering
    \includegraphics[width=\columnwidth ]{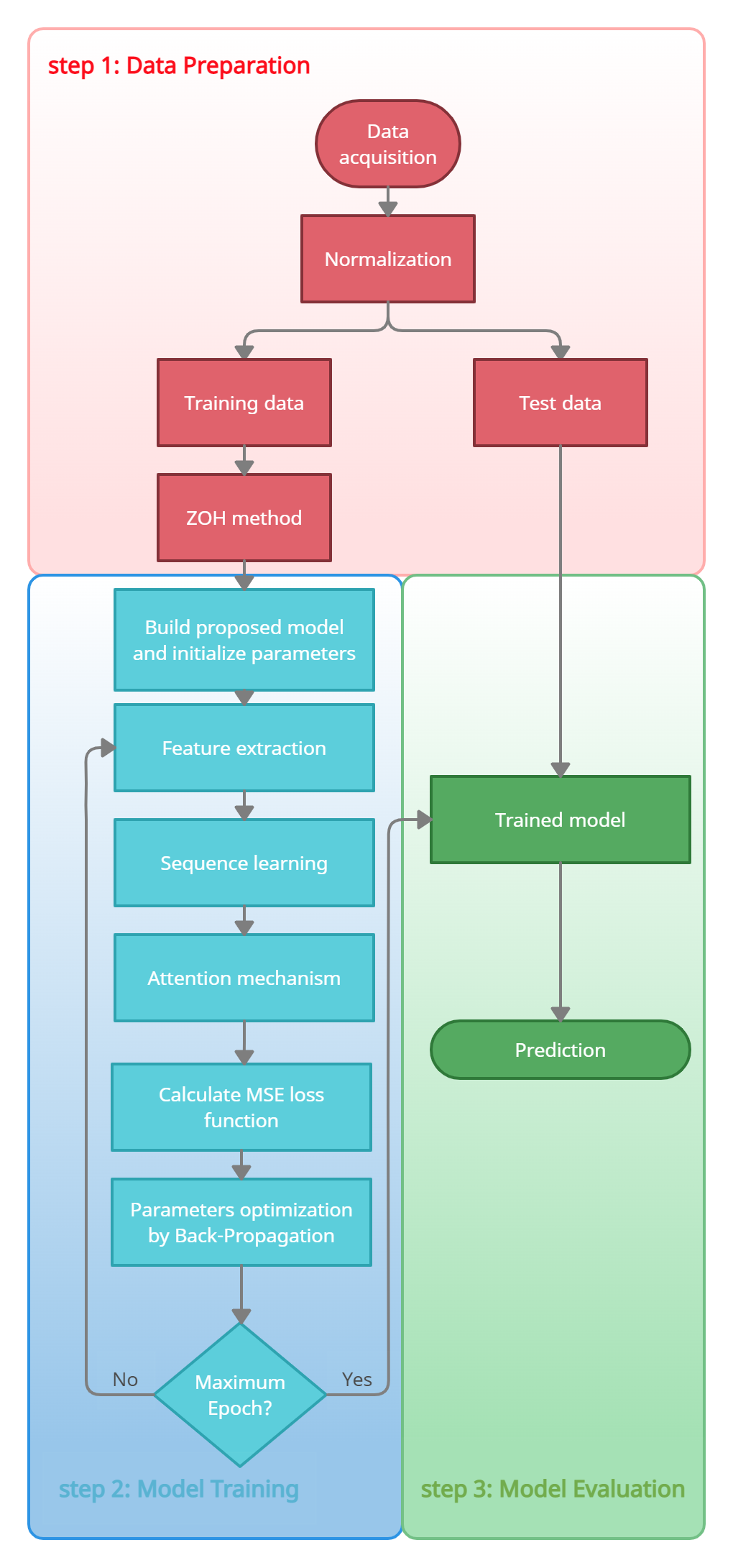}
    \caption{The flow chart of the proposed method}
    \label{flowchart}
    \end{figure}

In general, our proposed method in this paper contains three main phases: data preparation, model training, and model evaluation, which is shown in the Fig. \ref{flowchart}. The first step in the data preparation phase is to collect data from existing data sources. Subsequently, the collected data is normalized and divided into two parts of training and testing data. Training data is pre-processed using the ZOH technique.
The second step is to build and train the proposed model. In this step, first, the parameters of the constructed model are determined with the initial values. The spatial features of the input data are obtained through the feature extraction section and transferred to the layers of the sequence learning. The sequence learning part is used transferred spatial characterizes to model time series data. Finally, AM selects the output characteristics of the sequence learning part that have a stronger relationship with the prediction results.
The parameters are updated  in each epoch utilizing the backpropagation algorithm, and this training process will continue until the maximum epoch is achieved. The third step is model evaluation, in which the trained model is used to predict test data. Through these three steps, earthquake prediction is made using the proposed model.

\section{EXPERIMENT AND COMPARISONS}
\subsection{Division study area}
\begin{figure}[t]
    \centering
     \centering
    \includegraphics[width=\columnwidth ]{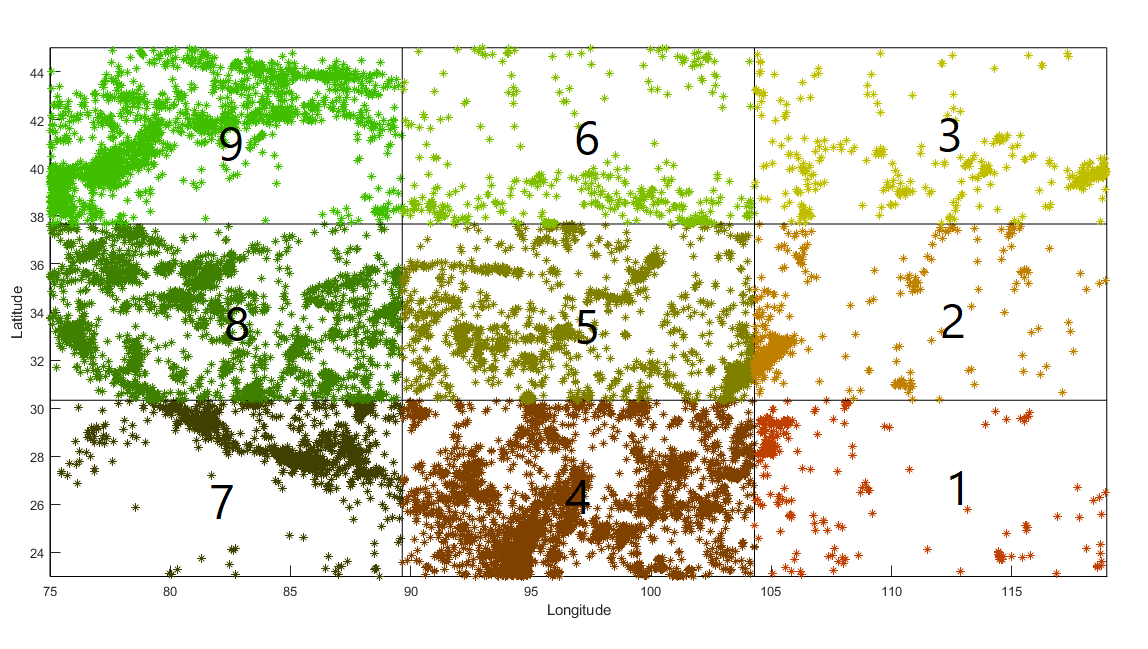}
   \caption{A visual representation of the division of China's nine study areas}
    \label{china}
    \end{figure}

 In this paper, mainland China has been chosen as the region of interest, situated in the southeast of the Eurasian plate. Mainland China is linked to the Siberia-Mongolia sub-plate, Philippine, and India plate; it is regarded as one of the most seismically active regions in the world. This country has already experienced many large and destructive earthquakes such as the 1966 Xingtai earthquake ($M_w$7.4), the 1975 Daguan earthquake ($M_w$7.1), the 2002 Jilin earthquake ($M_w$7.2), the 2008 Wenchuan earthquake ($M_w$8.0),  the 2013 Lushan earthquake($M_w$7.0), and the 2015 Gorkha earthquake($M_w$W7.8).
 Since 1949, more than 100 catastrophic earthquakes have occurred in mainland China. These earthquakes caused the death of more than 270,000 people, accounting for 54\% of the overall death toll from natural catastrophes in mainland China \cite{xi2017seismo}.
 Therefore, reliable and effective predictions in this area can help to reduce the damage and casualties caused by earthquakes.
 One of the aims of the earthquake prediction problem is to predict and identify regions where major earthquakes occur.
 In order to analyze and more accurately predict the range of location of the next earthquake, mainland China is divided into several smaller regions. However, the lack of enough and appropriate data makes earthquake prediction and model training in small areas challenging. The study area was divided into nine small areas to address this challenge. The latitude range from 23 to 45 degrees, and the longitude range from 75 to 119 degrees; the range of latitude and longitude is divided into three equal parts. Fig. \ref{china} depicts the nine study areas.

\subsection{Dataset and Data Preprocessing}
Typically, to characterize the main features of regional seismicity, the information of all earthquakes should be collected in the desired range. The database of earthquakes used in this paper has been obtained from the US Geological Survey (USGS) and National Seismological Center (NSC) websites. In order to produce a quality seismic catalog, any duplicate data from the rows are recognized and eliminated. The earthquake catalog consists of 11442 cases with a magnitude greater than or equal to 3.5, which occurred from January 15, 1966 to May 22, 2021. Each recorded earthquake contains vital information such as latitude and longitude, time of earthquakes, magnitude and depth of earthquakes, station number, etc. In addition to the importance of model learning structure, proper input and output adjustment also significantly impacts model learning performance.
Therefore, this study intends to investigate its impacts through two case studies with different input and output. In the present study, the input and output data in case study 1 is the number of earthquakes per month, and case study 2 is the maximum magnitude of earthquakes per month. Notice, when the time interval is one month is considered, there are 665 data samples.\\

Min-Max standardization \cite{moreira2021effects} was used to transforms the data to a specific range $[0,1]$. Furthermore, 80\% of the data is utilized for model training, with the remaining 20\% used for testing. Also, given that earthquakes are complex, nonlinear, and non-stationary time series, the existence of these zero values negatively impacts model training and makes prediction difficult. On the other hand, these values are not part of the system behavior, and by filling the zero values, the effect of earthquake complexity can be reduced and result in better predictions. Therefore, the ZOH technique is employed in training data to improve the generalizability and robustness of the proposed method, in which the zero values in each month are replaced with the last non-zero value of the previous month.

%


\subsection{Implementation details}
The selection of the model's hyper-parameters is crucial to the well-trained model. This study is adjusted most of the hyper-parameters through trial and error. For example, several optimization algorithms are compared including, stochastic gradient descent (SGD) \cite{njock2020evaluation}, RMSprop \cite{xu2021convergence} and Adam \cite{ding2020conditional}. According to the comparison results, Adam can enhance the suggested model's accuracy and is chosen as an optimization algorithm. Mean square error (MSE) is utilized as a loss function that can be backpropagated to update the weights and biases. The starting learning rate is adjusted to 0.001 and gradually decreased linearly to 0.0001 by the final epoch. This helps to keep a relatively stable pace during the learning process. The epoch number is set to be 150, and the batch size is 32. Each experiment is repeated ten times to limit the influence of random factors on network performance. Since each region has different geological characteristics and activities, separate training is necessary. Therefore, all nine regions use the same configurations and settings to train the network. The prediction results were obtained for each of the nine areas individually on the test dataset.

\subsection{Evaluation metrics}
In this study, the three evaluation criteria, such as root mean square error (RMSE), mean absolute error (MAE), r-squared ${R^2}$ are employed  to evaluate the performance of the proposed model and equitable comparisons with other models.
 The equations of RMSE, MAE, and $R^2$ are represented as follows:

\begin{equation}
\begin{aligned}
RMSE = \sqrt {\frac{1}{n}\sum\limits_{i = 1}^n {{{\left( {{y_i} - {{\overset{\lower0.5em\hbox{$\smash{\scriptscriptstyle\frown}$}}{y} }_i}} \right)}^2}} }  
\end{aligned}
\end{equation}

\begin{equation}
\begin{aligned}
MAE = \frac{1}{n}\sum\limits_{i = 1}^n {\left| {{y_i} - {{\overset{\lower0.5em\hbox{$\smash{\scriptscriptstyle\frown}$}}{y} }_i}} \right|} 
\end{aligned}
\end{equation} 

\begin{equation}
    \begin{aligned}
    {R^2} = 1 - \frac{{\sum\limits_{i = 1}^n {{{\left( {{{\overset{\lower0.5em\hbox{$\smash{\scriptscriptstyle\frown}$}}{y} }_i} - {y_i}} \right)}^2}} }}{{\sum\limits_{i = 1}^n {{{\left( {{{\bar y}_i} - {y_i}} \right)}^2}} }}
    \end{aligned}
\end{equation}
 In the aforementioned formulas, $n$ is the number of predicted data; the predictive value is ${{{\overset{\lower0.5em\hbox{$\smash{\scriptscriptstyle\frown}$}}{y} }_i}}$; the actual value is ${{y_i}}$, and ${\bar y}$ is the average of actual value. RMSE, MAE, R are the most common evaluation criteria for regression issues used in this study. In order to enhance the performance of the predicted model, RMSE, and MAE  should be minimized, while $R^2$ needs to be maximized. The $R^2$ metric indicates the strength of the linear regression between observed and predicted values. When $R^2$ equals one, the strongest linear relationship occurs. The RMSE and MAE metrics indicate the performance of the models. Both metrics can have a value ranging from zero to Infinite, with zero indicating the highest performance.

\subsection{Comparison method}
The models commonly utilized in shallow machine learning methods for earthquake prediction are SVM, MLP, DT, and RF. In deep learning neural networks, CNN, LSTM, and CNN-BiLSTM have satisfactory performance at prediction for time series data. Therefore, SVM, MLP, DT,  RF,  CNN, LSTM, and CNN-BiLSTM are chosen as the comparison methods for our proposed method. All Comparison models utilized a similar dataset and prediction process to the proposed method. In this article, comparison methods are divided into two major categories:
\\
\begin{itemize}
\item \textbf{Shallow machine learning:} SVM, DT, MLP, and RF are considered in the shallow machine learning category. These methods are proposed to evaluate the feature extraction ability of the proposed method compared to machine learning methods. 
\\
SVMs are machine learning algorithms are based on statistical learning theory. The main hyperparameters of the SVM model are set as follows: the type of kernel function is chosen Radial basis function (RBF) because it has fewer hyperparameters, which decreases the complexity of the model. The regulation factor (C), epsilon (e), and gamma (c) hyper-parameters are set to 1, 0.01, and 0.1, respectively.  
\\
RF is an efficient ensemble learning technique that employs averaging to boost prediction performance and avoid overfitting. It is extensively utilized in a variety of regression problems. In this experiment, the maximum depth of the trees and the number of trees in the forest are set to 9 and 100, respectively.
\\
A DT algorithm is a supervised learning approach that uses a tree structure to generate regression or classification models. The objective is to build a model that predicts the value of a target variable using simple decision rules inferred from data features. In this experiment, the maximum depth of the tree is chosen to be 10.

The MLP is one of the Neural Networks, which is often used in earthquake prediction studies \cite{mahmoudi2016predicting}. MLP models are composed of three primary layers: one input layer, some hidden layer, and one output layer. Their performance is determined by their structure, activation functions, and the manner connection weights between processing components are updated \cite{haykin1998neural}. Two hidden layers with 15 neurons in each layer are used in our case. The learning rate is set to 0.01. The epoch is 150, and the activation function is sigmoid.

\item \textbf{Deep learning:}  The deep learning category includes CNN, LSTM, and CNN-BiLSTM methods. The CNN method is related to the feature extraction block of the proposed model. The LSTM method is also similar to the proposed model's sequence learning block, except that LSTM is used instead of BiLSTM.
The CNN-BiLSTM approach is built with a combination of feature extraction block and sequence learning block.
\end{itemize}
In addition, all comparative methods without the ZOH technique are performed to show the effectiveness of the proposed ZOH technique in earthquake prediction.

\section{RESULT}

\subsection{Case 1: Number of earthquake}
The majority of research used seismic indicators, magnitude, depth, and geographical location of earthquakes as input and has only been able to predict the earthquake's time, location, and magnitude. To the best of our knowledge, no predictions concerning the number of earthquakes have been proposed. The number of earthquakes can be an essential factor in predicting an earthquake that can assist in portraying a more accurate picture of a region's seismicity. 
This case study is used with the same conditions and assumptions of the proposed method to predict the number of expected earthquakes in a month.
\\

The proposed method is compared with CNN-BiLSTM, LSTM, CNN, RF, MLP, DT, and SVM to verify the efficiency, superiority, and generalization ability.
Table \ref{number_earth} shows the results of the proposed method and comparison methods in predicting the number of earthquakes for nine different regions. 

\begin{table*}[]
\caption{Comparison of evaluation error indexes of the eight methods for number earthquake}
\label{number_earth}
\resizebox{\textwidth}{!}{%
\begin{tabular}{cccccccccc}
\hline
\multirow{2}{*}{Region}   & \multirow{2}{*}{Metric} & \multicolumn{8}{c}{Method}                                                             \\ \cline{3-10} 
                          &                         & SVM    & DT    & MLP   & RF    & CNN   & LSTM  & CNN-BiLSTM & \textbf{Proposed method} \\ \hline
\multirow{3}{*}{Region 1} & RMSE                    & 0.103  & 0.094 & 0.107   & 0.094 & 0.095 & 0.093 & 0.089      & 0.024                    \\
                          & MAE                     & 0.061  & 0.044 & 0.048 & 0.44  & 0.051 & 0.049 & 0.041      & 0.018                    \\
                          & ${R^2}$                 & -0.011 & 0.157  & 0.050  & 0.154  & 0.143  & 0.174  & 0.249       & 0.956                     \\ \hline
\multirow{3}{*}{Region 2} & RMSE                    & 0.125  & 0.094 & 0.145 & 0.128 & 0.083 & 0.073 & 0.069      & 0.064                    \\
                          & MAE                     & 0.086  & 0.108 & 0.092 & 0.058 & 0.049 & 0.042 & 0.037      & 0.031                    \\
                          & ${R^2}$                 & 0.604   & 0.660    & 0.432  & 0.584  & 0.826  & 0.864  & 0.882       & 0.915                     \\ \hline
\multirow{3}{*}{Region 3} & RMSE                    & 0.282  & 0.259 & 0.182 & 0.201 & 0.199 & 0.192 & 0.189      & 0.069                    \\
                          & MAE                     & 0.152  & 0.152 & 0.133 & 0.112 & 0.112 & 0.119 & 0.118      & 0.030                    \\
                          & ${R^2}$                 & -0.226  & -0.030 & 0.434  & 0.37  & 0.412  & 0.450  & 0.474       & 0.937                     \\ \hline
\multirow{3}{*}{Region 4} & RMSE                    & 0.146  & 0.145 & 0.138 & 0.139 & 0.130 & 0.128 & 0.100       & 0.096                    \\
                          & MAE                     & 0.088  & 0.084 & 0.072 & 0.077 & 0.069 & 0.065 & 0.066      & 0.062                    \\
                          & ${R^2}$                 & 0.011   & 0.138  & 0.263  & 0.214  & 0.432  & 0.495  & 0.613       & 0.697                     \\ \hline
\multirow{3}{*}{Region 5} & RMSE                    & 0.196  & 0.123 & 0.097 & 0.106 & 0.092 & 0.076   & 0.088        & 0.059                    \\
                          & MAE                     & 0.125  & 0.082 & 0.071 & 0.075 & 0.072 & 0.07  & 0.067      & 0.047                    \\
                          & ${R^2}$                 & -0.049  & 0.583  & 0.749  & 0.684  & 0.695  & 0.736  & 0.721       & 0.905                     \\ \hline
\multirow{3}{*}{Region 6} & RMSE                    & 0.229  & 0.197 & 0.174 & 0.192 & 0.195 & 0.196 & 0.183      & 0.088                    \\
                          & MAE                     & 0.142  & 0.125 & 0.011 & 0.126 & 0.14  & 0.132 & 0.125      & 0.06                     \\
                          & ${R^2}$                 & -0.347  & 0.019  & 0.213  & 0.059  & 0.054  & 0.096  & 0.154       & 0.805                     \\ \hline
\multirow{3}{*}{Region 7} & RMSE                    & 0.159  & 0.164 & 0.106 & 0.095 & 0.093 & 0.093 & 0.086      & 0.099                    \\
                          & MAE                     & 0.102  & 0.138 & 0.049 & 0.039 & 0.033 & 0.035 & 0.036      & 0.026                    \\
                          & ${R^2}$                 & 0.132   & 0.110  & 0.219  & 0.365  & 0.431  & 0.458  & 0.476       & 0.609                     \\ \hline
\multirow{3}{*}{Region 8} & RMSE                    & 0.141  & 0.141 & 0.134 & 0.141 & 0.130 & 0.131 & 0.115      & 0.092                    \\
                          & MAE                     & 0.078  & 0.074 & 0.067 & 0.075 & 0.058 & 0.056 & 0.062      & 0.06                     \\
                          & ${R^2}$                 & 0.075   & 0.182  & 0.264  & 0.180  & 0.392  & 0.385  & 0.426       & 0.669                     \\ \hline
\multirow{3}{*}{Region 9} & RMSE                    & 0.122  & 0.122 & 0.124 & 0.119 & 0.120  & 0.115 & 0.097      & 0.087                    \\
                          & MAE                     & 0.078  & 0.091 & 0.099 & 0.093 & 0.091 & 0.087 & 0.074      & 0.059                    \\
                          & ${R^2}$                 & 0.612   & 0.509  & 0.475  & 0.524  & 0.560  & 0.573  & 0.633       & 0.812                     \\ \hline
\end{tabular}%
}
\end{table*}

\begin{figure}[h]

    \centering
    \includegraphics[width=\columnwidth ]{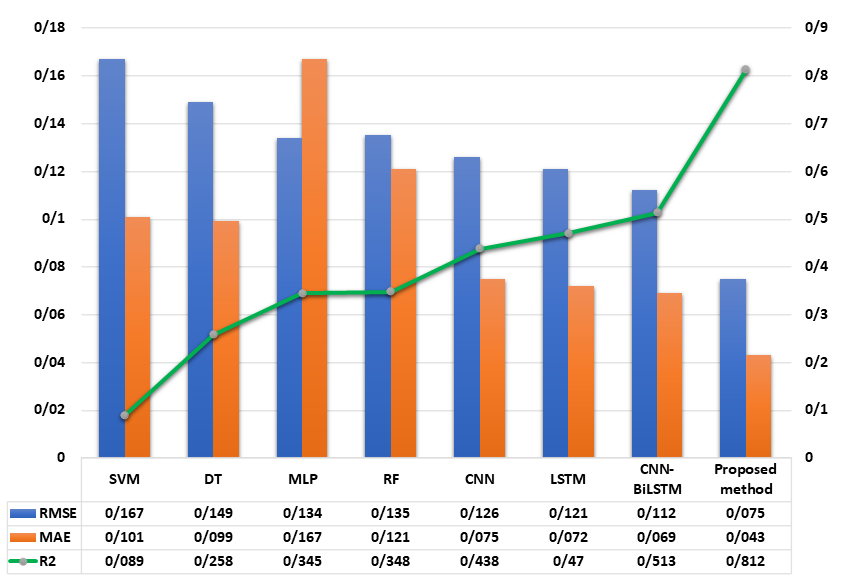}
    \caption{Comparison of the mean of the three evaluation metrics  between comparison methods and the proposed method for number of earthquake}
    \label{mean_number}
    \end{figure}

It is clear from the outcomes that the proposed method consistently achieves the best prediction performance in all regions, with the lowest RMSE and MAE values and the highest $R^2$ score. This shows that the proposed method is appropriately performed in predicting the number of earthquakes and its superior or competitive to other comparison methods.
For example, in region 1, the other comparative models present high prediction errors, while the proposed model improvements prediction results with the RMSE value 0.24, the MAE value 0.018, and $R^2$ value 0.956.
  \begin{figure*}[t]

    \centering
    \includegraphics[width=\textwidth]{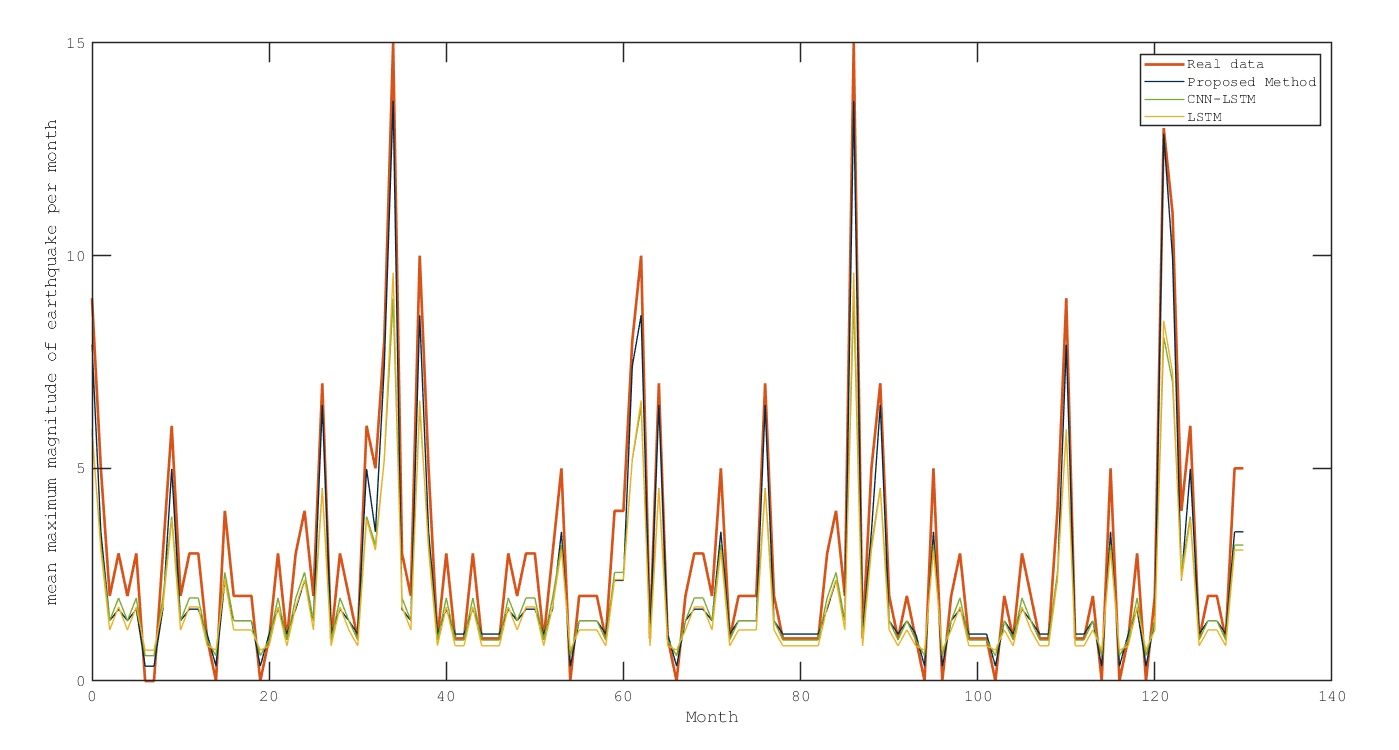}
    \caption{The comparison of proposed model and deep learning models for number of earthquake prediction in region 5.}
    \label{number5}
    \end{figure*}
    
In one of the most challenging regions in terms of $R^2$ metric, namely region 7, the proposed method has 0.133 and 0.5 better than the maximum and minimum $R^2$ values in comparison methods, respectively. When there is a significant seismic anomaly, this high performance illustrates the effectiveness of the proposed method.
  The reason for the better performance of the proposed method is that it has three key characteristics which can significantly improve prediction performance: using the ZOH technique as a suitable preprocessing, learning the spatial and temporal characteristics using the CNN-BiLSTM hybrid model, applying the idea of AM to considering the importance of different hidden states and more focus on the most important states. As a result, these three types of characteristics together make the proposed method superior on both ML-based and DL-based approaches.
  On the other hand, the SVM model has the worst prediction performance in all regions, with the highest RMSE and MAE values and lowest $R^2$ score. SVM method fails to catch the long-term correlation in the data series, despite its ability to solve nonlinear problems. 
  \\
 
Additionally, we also give the mean of evaluation metrics of each implemented method to demonstrate our proposed method's overall advantages various methods for the number of earthquakes in all regions.
  As can be seen Fig. \ref{mean_number}, there are clear disparities in the performance of various techniques. Specifically, the deep learning-based methods have a lower average error and more substantial predictive potential than shallow machine learning-based methods. This fact is that the shallow machine learning methods structure is relatively simple. While there is a nonlinear relationship between the earthquake data and the variables related to it. 
 Therefore, deep learning-based approaches are more adaptable in modeling complicated and nonlinear earthquake interactions and, by introducing several hidden layers, they can provide the capability of learning features at different levels of abstraction. Also between comparison models of deep learning , the LSTM model (RMSE = 0.121, MAE = 0.072,$R^2$= 0.47) performed better than the CNN model (RMSE = 0.126, MAE = 0.075, $R^2$=0.438). Because in time series prediction, the LSTM model outperformed the CNN model. This is likely due to the fact that the CNN model cannot capture the long-term dependency.
  The hybrid model CNN-BiLSTM model has the lowest error values and highest $R^2$ score compared with CNN and LSTM single. This is because that the hybrid model incorporates the advantages of both LSTM and CNN, with LSTM capturing temporal features and CNN capturing spatial features. The proposed ( RMSE = 0.075, MAE = 0.043, $R^2$ = 0.812) method performed better than the CNN-BiLSTM method for number of earthquake prediction. 
  Overall, our proposed model shows improvements of 3.07\% and 1.1\% beyond the CNN-BiLSTM model on RMSE, MAE, and $R^2$ metrics, respectively. This improvement is due to the effectiveness of the two proposed ZOH and AM schemes. In other words, this demonstrates that the introduction of AM and ZOH into the original CNN-BiLSTM model can fully explore the complex relationship and enhance the earthquake prediction performance.

   Fig. \ref{number5} displays the comparison of the proposed method and deep learning models for predicting the number of earthquakes in region 5. Similar outcomes can also be found in other regions.
  From Fig. \ref{number5}it can be seen, the proposed method is better fitting impact than other methods and, in most cases, closer to the observed values.

\subsection{Case 2: Maximum magnitude of earthquake}

\begin{table*}[t]
\caption{Comparison of evaluation error indexes of the eight methods for Maximum magnitude}
\label{max_earth}
\resizebox{\textwidth}{!}{%
\begin{tabular}{cccccccccc}
\hline
\multirow{2}{*}{Region}   & \multirow{2}{*}{Metric} & \multicolumn{8}{c}{Method}                                                              \\ \cline{3-10} 
                          &                         & SVM   & DT    & MLP    & RF    & CNN   & LSTM   & CNN-BiLSTM & \textbf{Proposed method} \\ \hline
\multirow{3}{*}{Region 1} & RMSE                    & 0.441 & 0.437 & 0.104 & 0.442 & 0.264      & 0.240  & 0.093      & 0.021                    \\
                          & MAE                     & 0.380 & 0.290 & 0.040  & 0.377 &    0.062   & 0.054  & 0.041      & 0.021                    \\
                          & ${R^2}$                 & -0.344 & -0.141 & 0.052   & -0.256 &    0.096   & 0.164 & 0.217       & 0.995                     \\ \hline
\multirow{3}{*}{Region 2} & RMSE                    & 0.438 & 0.234 & 0.821  & 0.194 & 0.128      & 0.109  & 0.073      & 0.042                    \\
                          & MAE                     & 0.210 & 0.207 & 0.191  & 0.168 &    0.175   & 0.149  & 0.041      & 0.040                    \\
                          & ${R^2}$                 & 0.502  & 0.722  & 0.107 & 0.654  & 0.592      & 0.680   & 0.867       & 0.991                     \\ \hline
\multirow{3}{*}{Region 3} & RMSE                    & 0.438 & 0.322 & 0.330  & 0.340 & 0.332      & 0.318  & 0.307      & 0.070                    \\
                          & MAE                     & 0.284 & 0.302 & 0.277  & 0.284 &    0.284   & 0.245  & 0.232      & 0.062                    \\
                          & ${R^2}$                 & -0.251 & 0.212  & 0.296  & 0.246  &  0.360     & 0.451   & 0.473       & 0.962                     \\ \hline
\multirow{3}{*}{Region 4} & RMSE                    & 0.518 & 0.446 & 0.428  & 0.416 &   0.405    & 0.370  & 0.342      & 0.132                    \\
                          & MAE                     & 0.3112 & 0.248 & 0.262  & 0.255 &    0.212   & 0.225  & 0.198      & 0.118                    \\
                          & ${R^2}$                 & -0.172 & 0.088  & 0.156   & 0.072  &   0.256    & 0.261   & 0.281      & 0.866                     \\ \hline
\multirow{3}{*}{Region 5} & RMSE                    & 0.186 & 0.165 & 0.203  &     0.151  &  0.149     & 0.149  & 0.147      & 0.029                    \\
                          & MAE                     & 0.161 & 0.143 & 0.185  & 0.134 &  0.136     & 0.134 & 0.135      & 0.039                   \\
                          & ${R^2}$                 & 0.252  & 0.424 & 0.158  & 0.502  &  0.497     & 0.509  & 0.545      & 0.971                    \\ \hline
\multirow{3}{*}{Region 6} & RMSE                    & 0.398 & 0.346 & 0.394  & 0.345 & 0.342 & 0.333       & 0.330      & 0.120                    \\
                          & MAE                     & 0.358 & 0.288 & 0.362  & 0.285 & 0.286 &   0.275     &   0.274    & 0.272                    \\
                          & ${R^2}$                 & -0.301 & 0.219  & -0.271  & 0.243  & 0.250 &   0.300     & 0.312      & 0.921                     \\ \hline
\multirow{3}{*}{Region 7} & RMSE                    & 0.205 & 0.190 & 0.209  & 0.187 &  0.185     & 0.184  & 0.181      & 0.128                    \\
                          & MAE                     & 0.186 & 0.171 & 0.198  & 0.170 &    0.167   & 0.167  & 0.165      & 0.082                    \\
                          & ${R^2}$                 & 0.256  & 0.421  & 0.302   & 0.445  &  0.501     & 0.508   & 0.550       & 0.809                     \\ \hline
\multirow{3}{*}{Region 8} & RMSE                    & 0.150 & 0.147 & 0.138  & 0.146 &    0.117   & 0.114  & 0.109      & 0.061                    \\
                          & MAE                     & 0.078 & 0.072 & 0.068  & 0.072 &       0.060& 0.059  & 0.048      & 0.024                    \\
                          & ${R^2}$                 & 0.145  & 0.178 &  0.221      & 0.174 &      0.421  & 0.450 & 0.506      & 0.914                     \\ \hline
\multirow{3}{*}{Region 9} & RMSE                    & 0.162 & 0.164 & 0.159  & 0.163 &    0.143   & 0.140  & 0.138      & 0.072                    \\
                          & MAE                     & 0.135 & 0.138 & 0.133  & 0.137 &   0.128    &      0.127  & 0.124      & 0.057                    \\
                          & ${R^2}$                 & 0.142  & 0.111  & 0.165   & 0.124  &   0.201    & 0.235   &    0.256        & 0.801                     \\ \hline
\end{tabular}%
}
\end{table*}

A major earthquake with a magnitude greater than five can cause great financial and human losses and cause great concern to society. The existence of these destructive earthquakes in the history of mainland China highlights the significance of the maximum earthquake magnitude of prediction in this earthquake-prone area. Also, a timely prediction of earthquake maximum magnitude can be an effective step in giving early earthquake warnings and reducing damage caused by its occurrence.
\\
Because major earthquakes occur in the region over a period of months, a prediction time span of one month is appropriate in this study \cite{asim2017earthquake}. However, there are also occasions where more than one major earthquake strikes the region in a month, but the algorithm only counts one. The overall percentage of months with major earthquakes is about 35\%, with the other months recording mainly low-level earthquakes.
\begin{figure}[h]

    \centering
    \includegraphics[width=\columnwidth ]{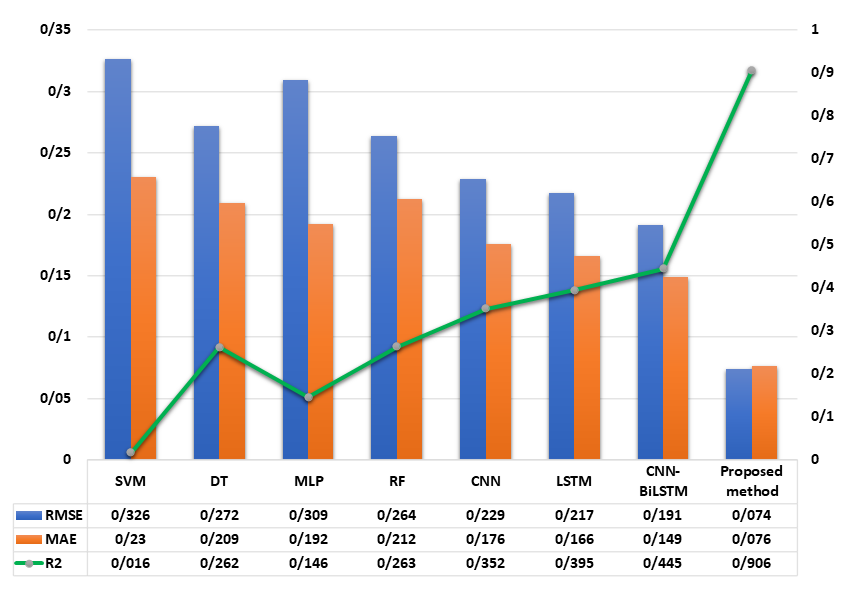}
    \caption{Comparison of the mean of the three evaluation metrics  between comparison methods and the proposed method for maximum magnitude}
    \label{mean_maximum}
    \end{figure}

\begin{figure*}[t]

    \centering
    \includegraphics[width=\textwidth ]{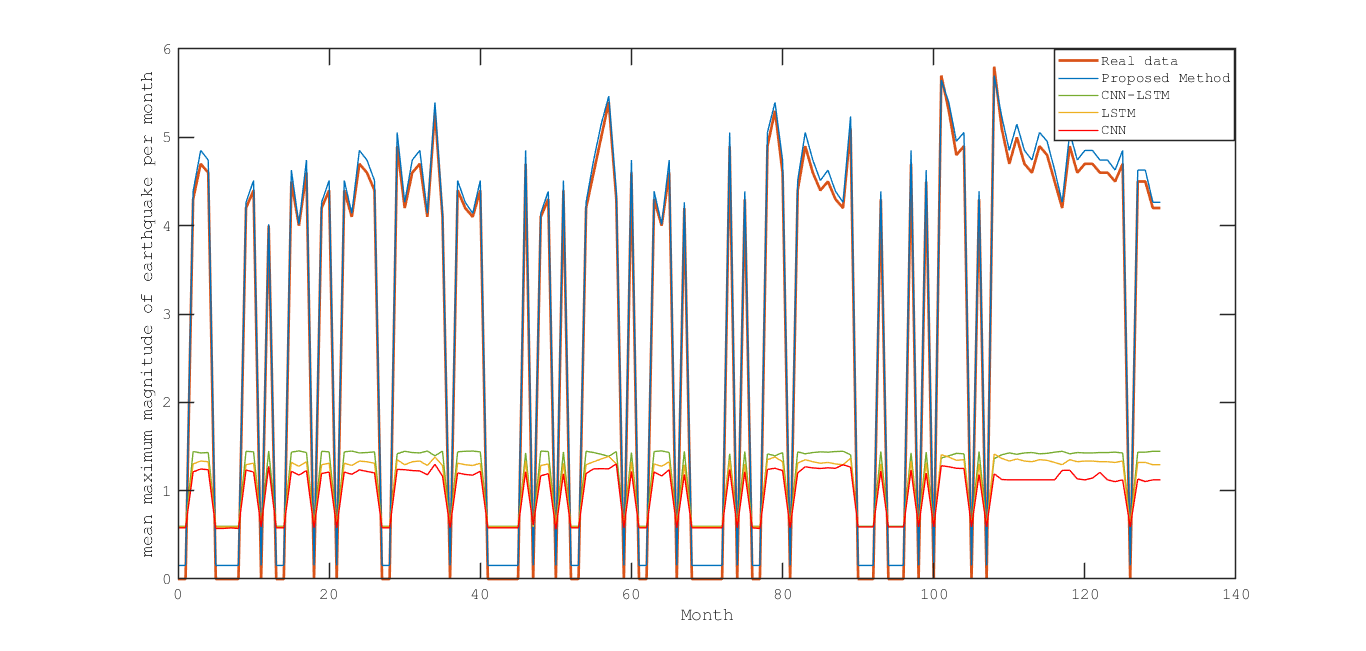}
    \caption{The comparison of proposed model and deep learning models for maximum magnitude prediction in region 1}
    \label{max_mag_1}
    \end{figure*}

In case study 2, we compare the performance of our proposed technique for maximum earthquake magnitude to various prediction methods. The input data contains the sequence data of the maximum magnitude of earthquakes every month for a span of time, and the output data includes the maximum magnitude of earthquakes predicted for the next month.

Table \ref{max_earth} shows the simulation results of the proposed model and comparison models for predicting the maximum earthquake magnitude in nine areas.
The results demonstrate that the proposed method consistently provides the strongest prediction performance in all areas, with the least RMSE, MAE values, and highest $R^2$ value.
This superiority is due to the three advantages of the proposed method: The spatial features and temporal correlations extracted by utilizing CNN-BiLSTM are considered to reflect the hidden features of the earthquake, The AM can highlight features that are important for earthquake prediction. The ZOH technique can help to reduce the effect of data’s zero-value on the network training process and reduce prediction error.
 For example, the proposed method shows improved results for regions 1 and 2 compared to other methods' performance with (0.995), and (0.991) of $R^2$. However, the $R^2$ value of proposed methods in these regions is 0.69 and 0.545 greater than the best $R^2$ value of the two categories, machine learning-based method and deep learning-based method.
 Moreover, the proposed method has shown the best prediction performance in one of the most difficult regions, such as region9 with $R^2$ of 0.801. while, the worst prediction performance in this region is related DT model with $R^2$= 0.111.  The proposed method has been able to improve the $R^2$ value is 0.69 and 0.545  higher than the best $R^2$ value of the two categories machine learning-based and deep learning-based.
 This high performance indicates the proposed method’s efficiency when there is a substantial distribution difference between train and test data.
 The difference in the probability and statistical distribution of the data is related to variances in seismic behavior in various regions caused by changes in the arrangement of its tectonic plates.

 To show the results more clearly in Table \ref{max_earth}, we calculate the values of the evaluation indices of all methods as averages and draw them in Fig. \ref{mean_maximum}. 
 Overall, all the deep learning models outperform the traditional models, and the proposed method has the best performance.
Take RMSE and MAE as an example. SVR, DT, MLP, RF, CNN, LSTM, CNN-BiLSTM obtains (0.326, 0.23), (0.272, 0.209), (0.309, 0.192), (0.264, 0.212), (0.229, 0.176), (0.217, 0.166) and (0.191, 0.149) at RMSE and MAE, respectively; while proposed method reduces RMSE and MAE to (0.074, 0.076), respectively. For $R^2$, the proposed method also achieves higher value (0.906) than the maximum values of the other comparison models. This minimum values of RMSE, MAE and maximum value of $R^2$ confirm the effectiveness of the ZOH and AM-based prediction method.
 Fig. \ref{max_mag_1} shows the comparison of the proposed model and deep learning models to predict the maximum magnitude of the earthquake in region 1. The proposed method is superior to other deep learning models on the upper and lower peak points, in terms of the trend shape and the fitting degree.

\section{conclusion}
Due to the nonlinearity and complexity nature of earthquakes data, this paper provides a new CNN-BiLSTM-AM approach and a novel and efficient general framework for earthquake prediction in terms of number and maximum magnitude. The number and maximum magnitude of earthquakes that occurred in each month over the past 50 years are considered the model's input features, which makes the model can completely extract useful information from the historical data. A new data processing technique called ZOH is presented to better train the network and lessen the prediction difficulty.
After data preprocessing, CNN is used to extract spatial characteristics. The features extracted by CNN are passed into BiLSTM. The BiLSTM is introduced to solve the data's long-term dependency, and the AM is used to highlight the BiLSTM output features that have a high contribution to the prediction results. Finally, the output of the AM is sent to the fully connected layers to obtain the final result. Compared to other shallow machine learning and deep learning approaches, the simulation results in two case studies reveal that the proposed method has the best performance.

\bibliographystyle{ieeetr}
\bibliography{main.bib}
\vfill

\end{document}